\documentclass[10pt,twocolumn,letterpaper]{article}

\usepackage{cvpr}              

\usepackage[dvipsnames]{xcolor}

\definecolor{cvprblue}{rgb}{0.21,0.49,0.74}
\usepackage[pagebackref,breaklinks,colorlinks,citecolor=cvprblue]{hyperref}

\def\paperID{9354} 
\def\confName{CVPR}
\def\confYear{2024}

\title{RustNeRF: Robust Neural Radiance Field with Low-Quality Images}

\author{
Mengfei Li\textsuperscript{1,2}, Ming Lu\textsuperscript{3}, Xiaofang Li\textsuperscript{4}, Shanghang Zhang\textsuperscript{2,*} \\
HKUST\textsuperscript{1}\quad Peking University\textsuperscript{2}\quad Intel Labs China\textsuperscript{3}\quad Southeast University\textsuperscript{4}
}

\begin{document}
\twocolumn[{%
\renewcommand\twocolumn[1][]{#1}%
\maketitle
\vspace{-10mm}
\begin{center}
    \centering
    \includegraphics[width=\linewidth]{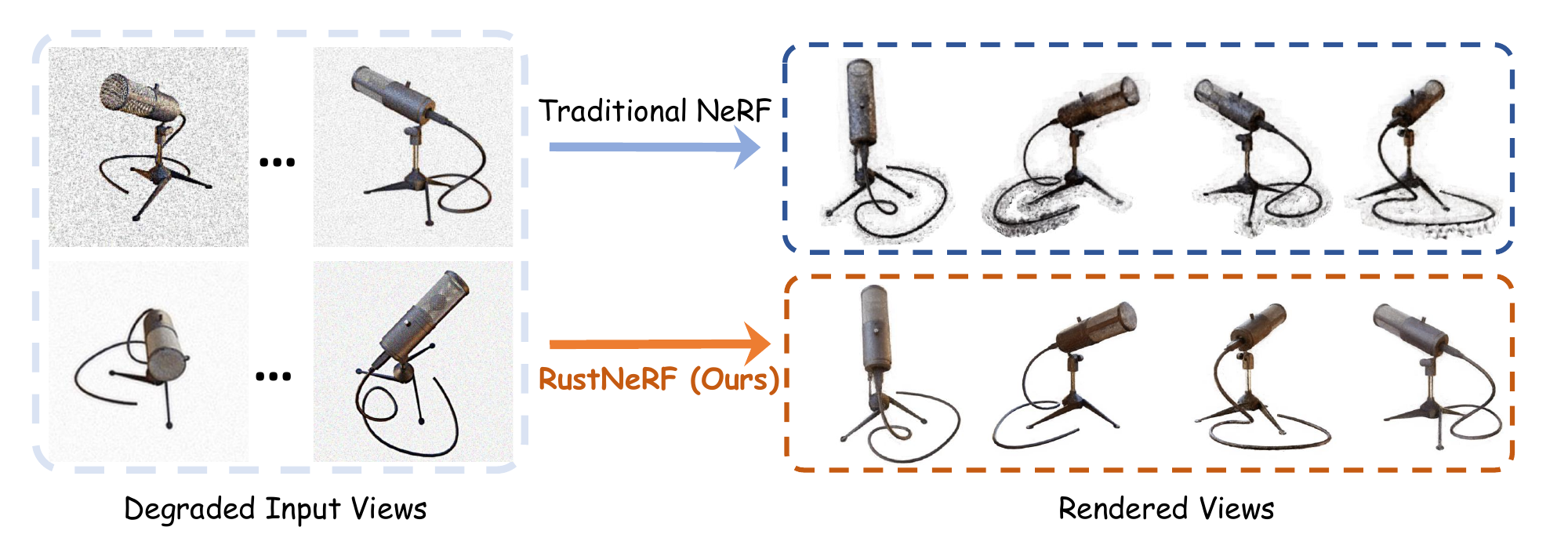}
    \vspace{-2em}
    \captionof{figure}{We present \textbf{RustNeRF}, a robust NeRF framework that can handle the degraded low-quality images. Traditional NeRF frameworks fail when encounter input views that are degraded for various reasons, and cannot get rid of artifacts when trained with these degraded images, while our RustNeRF can render high-fidelity results.}
    \label{fig:teaser}
\end{center}
}]
\maketitle


\begin{abstract}

Recent work on Neural Radiance Fields (NeRF) exploits multi-view 3D consistency, achieving impressive results in 3D scene modeling and high-fidelity novel-view synthesis. However, there are limitations. First, existing methods assume enough high-quality images are available for training the NeRF model, ignoring real-world image degradation. Second, previous methods struggle with ambiguity in the training set due to unmodeled inconsistencies among different views. In this work, we present RustNeRF for real-world high-quality NeRF. To improve NeRF's robustness under real-world inputs, we train a 3D-aware preprocessing network that incorporates real-world degradation modeling. We propose a novel implicit multi-view guidance to address information loss during image degradation and restoration. Extensive experiments demonstrate RustNeRF's advantages over existing approaches under real-world degradation. The code will be released.

\end{abstract}    
\section{Introduction}

Neural Radiance Fields (NeRF) have attracted great attention from the research community over the past few years. By learning the neural representation from sparse images of a complex scene, NeRF can render novel views of the scene. The pioneering work \cite{mildenhall2021nerf} uses Multi-Layer Perceptrons (MLPs) to regress the density and color of a scene from the input position. Volume rendering is then used to generate the rendered frame from the regressed density and color. In order to accelerate NeRFs, explicit neural representations \cite{muller2022instant,fridovich2022plenoxels,yu2021plenoctrees,sun2022direct} are proposed to replace the implicit neural representation of MLPs. Apart from this, there are a growing number of NeRF improvements emerged, e.g., extending NeRF to dynamic scenes \cite{pumarola2021d,park2021hypernerf}, training NeRF without camera poses \cite{lin2021barf,chen2022local}, reconstructing from few-shot samples \cite{yu2021pixelnerf,yang2023freenerf}, editing the scene with NeRF \cite{zheng2022editablenerf,yuan2022nerf}, learning NeRF for large-scale scenes \cite{tancik2022block,turki2022mega} and applying NeRF to human faces \cite{gafni2021dynamic,gao2022reconstructing}.

\begin{figure*}[ht]
    \centering
    \includegraphics[width=\linewidth]{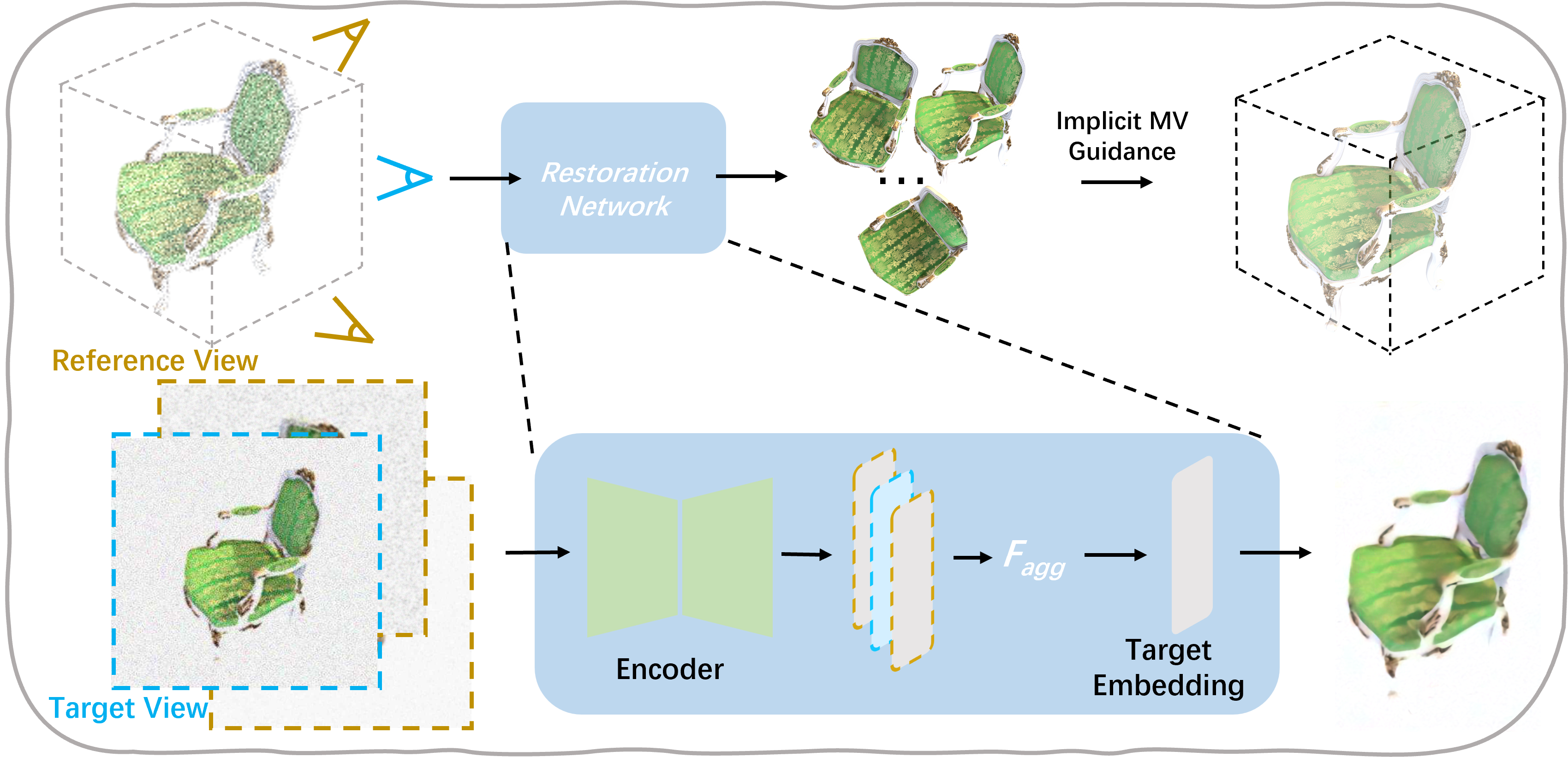}
    \caption{An overview of RustNeRF. Input views are first restored via restoration network. The restored views are used for supervising the training process of NeRF model. We further adopt implicit multi-view guidance to enhance the details via excavating the redundant information in multiple views.}
    \label{fig:single_image}
\end{figure*}


Although plenty of works are proposed, existing methods focus on generating higher resolution output from the input images, which overlook the quality of the training image-set. This problem is more significant when the users are using custom real-world datasets, where various types of degradations might occur. These degradations can be caused by the imaging system, ISP pipeline, compression and etc. The combinations of real-world degradations\cite{wang2021real} disastrously harm the fidelity of the custom data, leading to unsatisfactory novel views and hindering NeRF from wider applications. In addition, the degradation process might introduce ambiguity to the image-set, because the same position in the real world can be different in different views due to degradations. The optimization of NeRF will be even harder in face of this kind of ambiguity and inconsistency.

The real-world degradations are complicate to model. Previous methods \cite{cai2019toward, wang2021real} have explored performing image restoration and super-resolution under a single-image blind super-resolution scheme. A straightforward solution is adopting these methods to restore the degraded image set before training the NeRF model. However, experiments show that these methods are not 3D-awareness, hence are not able to restore the details in the original images. Moreover, inconsistency is 
aggravated since the images are restored independently. These problems makes the training end up with a sub-optimal NeRF model.

To solve the above limitations, we propose RustNeRF for real-world high-quality view synthesis. Specifically, we train a 3D-aware pre-processing network by incorporating real-world degradation modeling in the network. The pre-processing network gathers the information from the target view and views that are relevant to the target view and restore high-fidelity images with 3D-awareness. In this manner, we can enhance the pre-processing quality with multi-view degraded images. As for the real-world degradation modeling, we use the high-order degradation modeling process following existing methods \cite{wang2021real,chan2022investigating}. The 3D-aware pre-processing network can effectively reduce the exaggerated artifacts of the neural field under real-world degradations.



To the best of our knowledge, we are the first work to exploit NeRF with low-quality real-world images. Our main contributions are summarized as follows:

\begin{itemize}

\item We propose RustNeRF, a novel neural radiance field framework that robustly handle the degraded low-quality real-world images.

\item We design a novel 3D-aware restoration network for the neural radiance field, reducing the exaggerated artifacts under real-world degradations.


\item We excavate the redundancy information between multiple views to further restore the details of the scene via novel implicit multi-view guidance.

\item We conduct detailed experiments to demonstrate the advantages of our method against the existing approaches.
\end{itemize}

\section{Related Work}

\paragraph{Neural Radiance Fields.}

The neural Radiance Field (NeRF) is one of the most important 3D representations and has gained much attention from the community. The pioneering work \cite{mildenhall2021nerf} learns an MLP to regress the color and density of a 3D point. Then they use volume rendering to render the pixel color based on the regressed colors and densities of 3D points along the corresponding ray. After training, NeRF \cite{mildenhall2021nerf} can render high-fidelity arbitrary novel views. To accelerate the vanilla NeRF, many methods are proposed to replace the implicit MLP representation with explicit representations. DVGO \cite{sun2022direct} presents a representation consisting of a density voxel grid for scene geometry and a feature voxel grid with a shallow network for complex view-dependent appearance. DVGO can achieve competitive quality and converges rapidly from scratch. PlenOctree \cite{yu2021plenoctrees} pre-tabulates the NeRF into a PlenOctree for real-time rendering. They also train NeRFs to predict a spherical harmonic representation of radiance, removing the viewing direction as an input. Plenoxels \cite{fridovich2022plenoxels} represents a scene as a sparse 3D grid with spherical harmonics and optimizes this representation via gradient methods and regularization. Instant-NGP \cite{muller2022instant} proposes a novel explicit representation consisting of a small neural network augmented by a multiresolution hash table of trainable feature vectors. There are also plenty of works extending NeRF to various application scenarios such as scalable large scene \cite{tancik2022block,turki2022mega}, 3D human face \cite{gafni2021dynamic,gao2022reconstructing}, 3D human body \cite{zhao2022human,peng2021animatable}, and few-shot reconstruction \cite{yang2023freenerf,yu2021pixelnerf}. However, these methods struggle to generate higher-resolution output from the input images, which limits their practical applications. Most recently, NeRF-SR \cite{wang2022nerf} extends NeRF to create HR novel views from LR inputs by a supersampling strategy that splits a pixel into grid sub-pixels. However, there are still several limitations when applying NeRF-SR to real-world images. In this work, we present RustNeRF for real-world high-quality NeRF.

\paragraph{Image Restoration.}
Image restoration is a crucial problem in computer vision and lots of methods have been proposed for different applications. Traditional methods explore the priors of natural images to restore the degraded images. For instance, \cite{he2010single} proposes the dark channel prior that most local patches in haze-free outdoor images contain some pixels which have very low intensities in at least one color channel, to recover the high-quality haze-free images. \cite{he2012statistics} observes that the statistics of the patch-matching offsets are sparsely distributed. They use this prior to filling the missing region by combining a stack of shifted images via optimization. Deep image prior \cite{ulyanov2018deep} shows that the structure of a generator network is sufficient to capture a lot of image statistics prior. DNN-based methods learn the restoration models on large-scale datasets. For example, DnCNN \cite{zhang2017beyond} uses the residual learning strategy to tackle several general image restoration tasks such as Gaussian denoising, single image super-resolution, and JPEG image deblocking. Recent methods \cite{liang2021swinir,zamir2022restormer,wang2022uformer} show impressive restoration performance based on the transformer models. Although these methods can achieve pleasing results, they cannot well handle real-world degradations.

\paragraph{Real-World Degradations.} 
The key to real-world image restoration is real-world degradation modeling. Real-ESRGAN \cite{wang2021real} introduces a high-order degradation modeling process along with the common ringing and overshoot artifacts. HGGT \cite{chen2023human} constructs a human-guided GT image dataset with both positive and negative samples and trains the real-world restoration models with the dataset. AnimeSR \cite{wu2022animesr} proposes to learn the animation degradation from real low-quality inputs and incorporates the learned ones into the degradation generation pipeline. VQD-SR \cite{tuo2023learning} decomposes the local details from the global structure and collects a rich-content dataset for extracting the priors. NeRFLiX \cite{zhou2023nerflix} designs a NeRF-style degradation modeling approach and constructs large-scale training data to remove NeRF-native rendering artifacts. Different from our work, NeRFLiX focus on utilizing high-quality image-set to refine the degraded novel views caused by NeRF-style degradation during training, while our work focus on dealing with the horrible artifacts of NeRF introduced by the degraded image-set under real-world degradations. 
\section{Preliminaries}

\subsection{Neural Radiance Fields}

NeRF uses MLP for implicit representation of 3D scene, mapping 3D point $\boldsymbol{x}_{k}$ and view direction $\boldsymbol{d}$ in space into volume density $\sigma_{k}$ and color $\boldsymbol{c}_{k}$.

\begin{equation}
        (\sigma_{k},\boldsymbol{c}_{k}) = f(\boldsymbol{x}_{k}, \boldsymbol{d})
\end{equation}

Given a pixel on an image, a ray $\boldsymbol{r}$ is casted to the pixel. Assume $K$ 3D points on the ray are sampled with interval $\delta$. These $K$ points are summed via volume rendering to get the pixel color:

\begin{equation}
	\boldsymbol{\widetilde{C}}(\boldsymbol{r})=\sum_{k=1}^{K}T_{k}(1-{\rm exp}(-\sigma_{k}\delta_{k}))\boldsymbol{c}_{k}, T_{k}={\rm exp}(-\sum_{j=1}^{k-1}\sigma_{j}\delta_{j})
\end{equation}

After obtaining the rendered image, NeRF calculates the reconstruction error between the rendered image $\hat{I}$ and the input image $I$ to achieve optimization.

\begin{equation}
	\mathcal{L}_{rec}=\sum||\hat{I}-I||^{2}_{2}
\end{equation}





\subsection{Real-World Degradation in NeRF}

Previous methods simply assumes that high-quality image-set is available. However, in practical application, the quality of the image-set is not guaranteed. When users use NeRF variants in real-world data, the image-set is likely to be degraded due to many reasons\cite{wang2021real}. 

To formalize this problem, we first assume that the image-set containing $N$ images of the same scene $\mathcal{I}=\{I_1, I_1, ..., I_N\}$ is degraded similarly, because in most cases the images from $\mathcal{I}$ are taken from the same device and undergo the same processing.

\begin{equation}\label{eq:degrade_basic}
    \tilde{\mathcal{I}} = \Phi(\mathcal{I}) = \{\phi_1(I_1), \phi_2(I_2), ..., \phi_n(I_n)\}
\end{equation}
where $\Phi$ represents the real-world degradation, $\phi$ denotes the degrade function for every single image, while image $I_i$ denotes a sample in the image-set $\mathcal{I}$.

In this situation, the training process of NeRF is likely to fall into sub-optimum. The output novel views might suffer from severe artifacts and low-quality. Thus, how to train a NeRF model with this kind of degraded image-set becomes a vital question towards wider practical application of NeRF. 

Although it's often hard to estimate the degradation process $\phi$, there are different degradation models and methods to estimate the degradation process. Modeling these might be helpful in certain circumstances. However, real-world degradations are far more complex and harder to be estimated as the degradation can be a combination of different degradations, and the degradation process is unknown. We point out that the uncertainty of the degradation process makes the scene restoration task a blind restoration problem, making the restoration even harder. Hence, we need to design a sophisticated way to restore the scene, keeping the scene's high fidelity and consistency.
\section{Method}




The image quality of the training set is important in training a NeRF. It is hard to reconstruct a complex scene well with degraded images of low quality. However, restoring a degraded image with an unknown and complex degradation process is a challenging task, which is called blind image restoration. In our task, how to reserve the multi-view consistency in the original image-set after restoration is even harder. In this paper, we design a 3D-aware scene restoration strategy to deal with the degraded real-world scene images, while reserving the multi-view consistency.

The main challenge of the task is lacking proper dataset. Previous datasets widely used by NeRF methods like Blender\cite{mildenhall2019local}, LLFF\cite{chen2022local, mildenhall2021nerf}, BlendedMVS\cite{yao2020blendedmvs}, Tanks and Temples\cite{Knapitsch2017} focus on specific scenes. However, they are all high-quality datasets and thus limit the scale of the data. To obtain large-scale data for training a 3D-aware restoration network, we leverage the characteristics of the video dataset, as consecutive frames in the video dataset are naturally multiple views of a scene, which is similar to the dataset used in NeRF. Although these datasets are not designed for restoration task, we can stimulate the real-world degradations in a scene and synthesize degraded data for training.

\subsection{Degradation Parameterization}

Real-world degradation is a very complex process, as the images might be influenced by many different unknown processing procedure, which is relavant to the imaging system, ISP pipeline and even software compression or resizing. Previous method\cite{wang2021real} proposed to use the combination of several classical degradations to simulate the real-world degradation, providing a good approximation for single image real-world. As the method is designed of single image, we further extend this scheme to the scenarios that are appropriate for NeRF.

We can formalize the scene degradation process as follow.

Firstly, every single image $I$ is degraded by a degradation process:
\begin{equation}
\phi(I) = \rho_1\circ\rho_2\circ\cdots\circ\rho_l(I;\Theta)    
\end{equation}
where $l$ is a number of classical degradation process and $\Theta=\{\theta_1, \theta_2,...,\theta_l\}$ is the parameters set for $\phi$. Each degradation process $\rho_i$ is parameterized by $\theta_i$. $\rho$ refers to classical degradation process.

Note that the parameter $\theta$ of each process is relevant to the type of the degradation. For example, we use the strength of noise to parameterize the degradation of adding noise and use resize scale to parameterize the resizing process. The parameters are sampled from a designed parameter space $\Omega$ to simulate the real-world degradation.

Based on the single image degradation parameterization, Eq.\ref{eq:degrade_basic} can be rewritten as:

\begin{equation}
    \tilde{\mathcal{I}} = \Phi(\mathcal{I};\Omega) = \{\phi_1(I_1;\Theta_1), ..., \phi_n(I_n;\Theta_n)\}
\end{equation}
where $\Theta_i \in \Omega$.

The images of the same scene are commonly taken with the same device and undergo very similar data processing. Inspired by this, we can derive the scene degradation as the following form.

\begin{equation}
    \tilde{\mathcal{I}} = \Phi(\mathcal{I};\Omega) = \{\phi(I_1;\Theta_1), ..., \phi(I_n;\Theta_n)\}
\end{equation}

Considering that in practical scenarios, the lighting condition and the status of the imaging system won't change significantly, the pipeline of Imaging and ISP should degrade the data with very similar pattern, i.e. similar degrade kernel and noise distribution. To this end, we can assume that the parameters for different degradation remain unchanged in the same scene, which yields the following equation.

\begin{equation}
    \tilde{\mathcal{I}} = \Phi(\mathcal{I};\Omega) = \{\phi(I_i;\Theta)|\Theta\in\Omega,i=\{1,2,...,n\}\}
\end{equation}

With our real-world degradation parameterization, the degradation process is specific to different scene. For more details about the design of the degradation model, we kindly suggest readers referring to our supplementary materials.

\subsection{Scene Restoration}

Although recently many work have studied restoring images in single-image or video blind restoration scheme, these methods are not appropriate in our task. Single image restoration lacks the ability of modeling 3D scene, while video restoration exploits the relationship between dense consecutive frames, which is not the case for NeRF that uses sparse image set to train.

Taking above problems into consideration, we use the scene degradation parameterization to synthesize a dataset from a video dataset, as frames in the same clip that are degraded with the same parameters are naturally a 3D scene.

First, we use a visual encoder $f_{enc}$ to encode each degraded image $\tilde{I_i}$ and get the image feature $E_i$.

\begin{equation}
    E_i=f_{enc}(\tilde{I}_i)=f_{enc}(\phi(I_i;\Theta))
\end{equation}

Intuitively, once the degradation of a scene is parameterized, we can use all the images in the scene to estimate the degradation process and restore each image $\tilde{I}_i$ to $\hat{I}_i$ with a restoration network $f_{res}$ as follow: 

\begin{equation}
    \hat{I}_i = f_{res}(E_i;f_{enc}(\tilde{\mathcal{I}}))
\end{equation}

But in reality, this is not practical as a scene usually contains hundreds of images. It is computationally-expensive to utilize all the data. Besides, our key insight is that, the information missing in an image might be available in the neighbor frames. Hence, if the reference view has little content in common with the target view, the information in this reference view is redundant. So we propose a 3D-aware restoration network, selecting k relevant views with view selection function $\kappa$ as reference to restore the target view. This can be formulated as:

\begin{equation}
    \hat{I}_i \approx f_{res}(E_i;f_{enc}(\kappa(\tilde{I_i})))
\end{equation}

We use the 3D-aware restoration network to restore the degraded frame and predict the high-quality frame $\hat{I_0}$ for NeRF model training. We restore every frame in the training set before training NeRF model.

The whole restoration network is trained on a synthesized dataset based on training set of LLFF\cite{mildenhall2019local} and Vimeo90K \cite{xue2019video}. For every clip, we adopt the same degradation process with parameter set $\Theta$, and we sample a new parameter set $\Theta$ in the degradation space $\Omega$ when sampling a new clip. We suggest readers referring to our supplementary material for more details about the architecture of the network, the information about the dataset, the training and inference process.




\subsection{Implicit Multi-View Guidance}


\begin{figure}[h]
    \includegraphics[width=\linewidth]{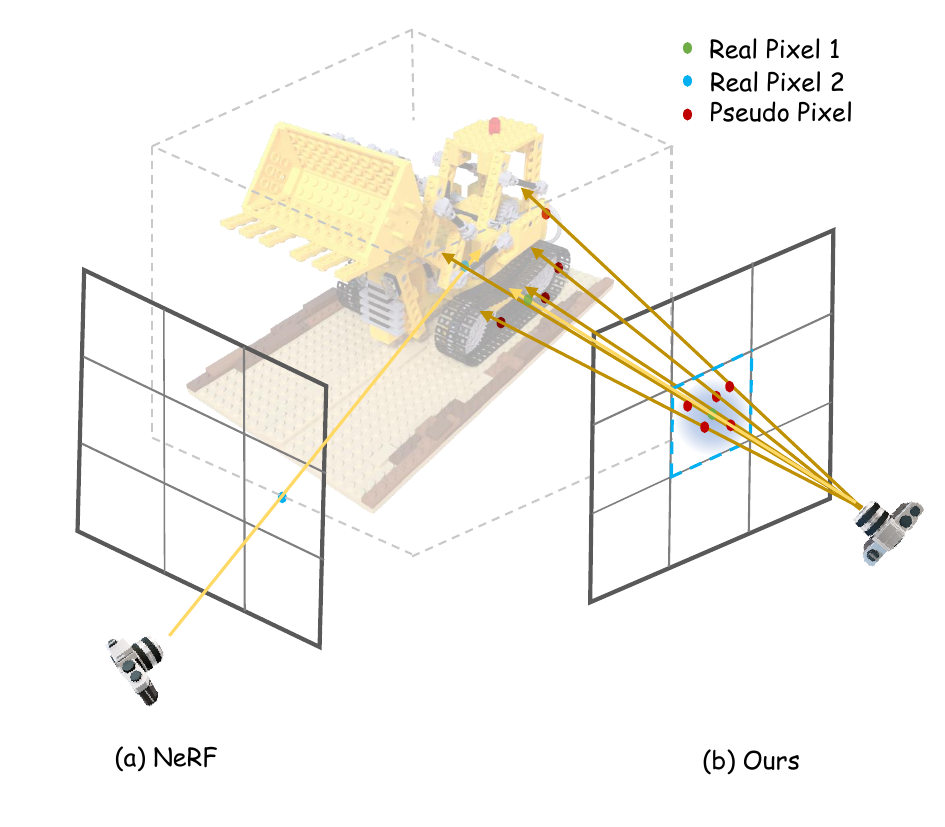}
    \caption{In (a) traditional NeRF training procedure, only the center of each pixel(Real Pixel) will be sampled, thus the whole pixel is supervised by such single inaccurate value. We propose (b) to cast multiple rays(Pseudo Pixels) inside the pixel and calculate the pixel value via weighted sum so as to excavate the supervision signal from other views. The pseudo pixel and real pixel are correspond to the same point in the scene, which illustrates our insight. Note that the distribution of rays in the figure is exaggerated.}
    \label{fig:supersampling}
\end{figure}

\begin{table*}[h]
    \centering
    \begin{tabular}{c|cccc|cccc}
    \toprule
       & \multicolumn{4}{c|}{Blender} &
           \multicolumn{4}{c}{LLFF} \\
    Method & $PSNR\uparrow$ & $SSIM\uparrow$ & $LPIPS\downarrow$ & $time\downarrow$ &
           $PSNR\uparrow$ & $SSIM\uparrow$ & $LPIPS\downarrow$ & $time\downarrow$ \\
    \midrule
    DVGO\cite{sun2022direct} & 21.05 & 0.824 & 0.253 & 499s & 15.62 & 0.484 & 0.685 & 359s\\
    Instant-NGP\cite{muller2022instant} & 22.72 & 0.731 & 0.494 & 986s & 20.84 & 0.623 & 0.539 & 1216s \\
    Ours(DVGO) & 24.07 & \textbf{0.850} & 0.229 & 639s & 16.31 & 0.500 & 0.660 & 513s\\
    Ours(Instant-NGP) & \textbf{24.40} & 0.841 & \textbf{0.201} & 1389s & \textbf{21.65} & \textbf{0.637} & \textbf{0.39} & 2707s \\
    \bottomrule 
    \end{tabular}
    \caption{The main quantitative experiment results of our methods and baseline methods on famous NeRF dataset Blender and LLFF. Our method gains significant improvement on this task, achieving the state-of-the art.}
    \label{tab:main results}
\end{table*}

Degradation naturally introduce severe artifact and ambiguity to the image-set. Although we adopt 3D-aware restoration network to reduce the loss of information, this problem still can not be totally solved. However, considering that degradation in the scene is a random process, even if a certain part of an image suffer from information loss and fails to be restored, relevant information still can be excavated from other views.

As the images are drastically degraded before restoration, the images still suffer from loss in details. This is because that the value of these pixels related to details are inaccurate. As a result, using these values to supervise the corresponding region in the image may lead to sub-optimum. To utilize the information in other views, instead of shooting single ray to the center of the pixel, we use a different ray distribution inside the pixel. We cast multiple rays to the pseudo pixels inside the pixel and calculate the real pixel value according to the rendered values of these pseudo pixels.

Through introducing these pseudo pixels inside the pixel, some rays will gather information from other views, as the rays might eventually stop at different points in the realistic scene rather than stopping at the original position as the ray going through real pixel does.

Then the question is how to aggregate these values to get the real pixel value. Instead of directly calculate the average value of these pseudo pixel values, we calculate the weighted sum of the the pseudo pixel values. Empirically, the weights should be relevant to the distance between the pseudo pixel and the real pixel. Thus we adopt a normal distribution. All weights $w_i$ follow a bivariate normal distribution centering at the coordinate of the real pixel. The covariance matrix $\sum$ is a hyper-parameter. Assume that the rendered pseudo pixel value is denoted as $\tilde{C}$, the real pixel value $\hat{C}$ can be calculated as:

\begin{equation}
    \hat{C}_{ij}=\sum_{k=1}^{p}w_k\tilde{C}^k_{ij}
\end{equation}

Finally, the loss function that uses implicit multi-view guidance can be fomalized as:

\begin{equation}
    \mathcal{L}_{MSE}=\sum_{i=1}^{n} |C(i, j) - \hat{C}(i,j)|
\end{equation}

\paragraph{Training Details}

Intuitively, at the early stage of the training, we don't have to apply implicit multi-view guidance as it introduces unnecessary training cost, because we cast multiple rays to each pixel. At the early stage, the NeRF model converges quickly to fit the coarse geometry of the scene, which does not need such complex training strategy. Instead, we propose to train NeRF model in a coarse-to-fine strategy. To be more specific, at the early stage of training, we adopt traditional NeRF training manner, and enable implicit multi-view guidance at the fine stage.

\paragraph{Quadtree Acceleration}

Intuitively, NeRF model can easily learn low-frequency information, which refers to areas with similar colors in larger areas, while it is more difficult for the model to learn high-frequency information, which includes edges and details. 

Information loss is likely to happen at the regions that contain more detail. While most regions of an image are composed of simple patterns, which are easy for restoration and reconstrcuted by NeRF model. Hence, we do not need to perform implicit multi-view guidance in these regions to prevent introducing too much unnecessary computation. We adopt the quadtree acceleration technique\cite{zhang2023fast} as we find that this method is especially appropriate for our implicit multi-view guidance.

\section{Experiments}

\begin{figure*}[t!]
    \centering
    \includegraphics[width=\textwidth]{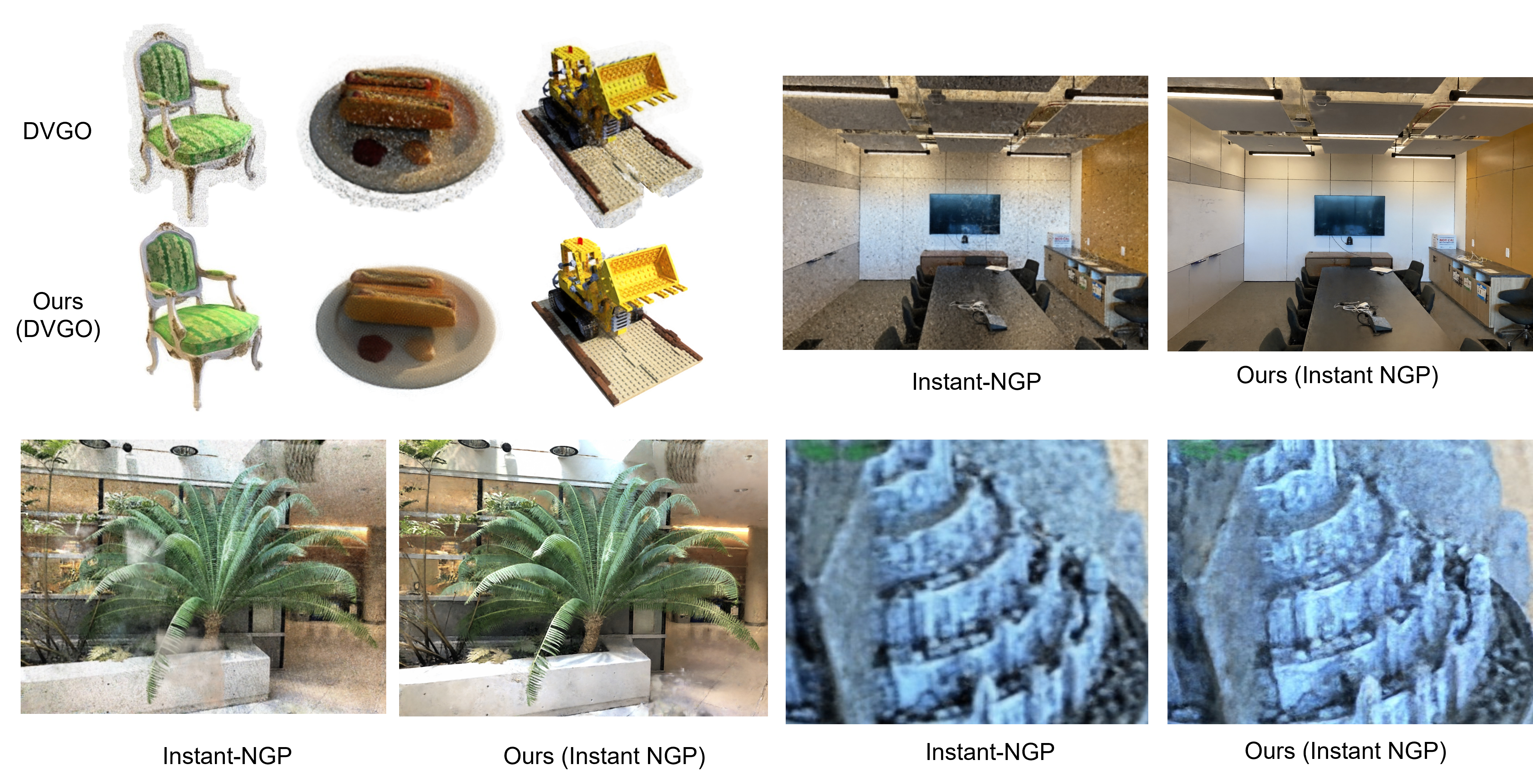}
  



    \caption{Qualitative results comparison of our method with baseline method. These scenes are taken from Chair(Blender), Hotdog(Blender), Fortress(LLFF) and Room(LLFF) respectively.}
    \label{fig:qualitative}
\end{figure*}

\subsection{Implementation Details}

When training the restoration network, we adopt horizontal flipping as data augmentation. We set the batch size to 8 and train the network for 300K iterations. We decay the learning rate from 5e-4 to 0 with a cosine annealing strategy. Adam\cite{kingma2014adam} optimizer is adopted, with $\beta_1=0.9,\beta_2=0.999$. We train the restoration network with 8 NVIDIA V100 GPU with 32GB VRAM.

We choose well-known DVGO\cite{sun2022direct} and Instant-NGP\cite{muller2022instant} as our base NeRF model. Our code is based on the official code of DVGO and a PyTorch implementation of Instant-NGP\footnote{We use the PyTorch implementation of Instant-NGP available at https://github.com/kwea123/ngp\_pl}. Training recipe is demonstrated in supplementary materials in detail.


\begin{table}[h] 
	\centering
        \begin{tabular}{c|c|c|c}
        \toprule
           Method & PSNR$\uparrow$ & SSIM$\uparrow$ & LPIPS$\downarrow$ \\
        \midrule
        w/o restoration & 23.69 & 0.839 & 0.233 \\
        Real-ESRGAN & 24.18 & 0.853 & 0.181 \\
        single-view & 23.97 & 0.842 & 0.195 \\
        multi-view & 24.42 & 0.852 & 0.182 \\
        multi-view(VS) & 24.90 & 0.851 & 0.187 \\
        \bottomrule
		\end{tabular}
        \caption{Quantitative results about different preprocessing method on Blender dataset.}
		\label{tab:ablation_pre_blender}
\end{table}

\begin{table}[h] 
	\centering
        \begin{tabular}{c|c|c|c}
        \toprule
           Method & PSNR$\uparrow$ & SSIM$\uparrow$ & LPIPS$\downarrow$ \\
        \midrule
        w/o restoration & 21.68 & 0.650 & 0.487 \\
        Real-ESRGAN & 20.77 & 0.652 & 0.479 \\
        single-view & 21.61 & 0.657 & 0.472\\
        multi-view & 21.46 & 0.656 & 0.473 \\
        multi-view(VS) & 22.01 & 0.657 & 0.472\\
        \bottomrule
		\end{tabular}
        \caption{Quantitative results about different preprocessing method on LLFF dataset.}
		\label{tab:ablation_llff}
\end{table}

\subsection{Datasets and Metrics}

We conduct our experiments on two famous datasets for novel view synthesis. These datasets cover synthesis data and realistic data, which helps to demonstrate the generalizability of our method.

$\textbf{LLFF}$. LLFF\cite{chen2022local, mildenhall2021nerf} consists of eight real scenes, each containing 20 to 62 input images. We adopt a resolution of 2016$\times$1512 for experiments.
 
$\textbf{Blender}$. The Realistic Synthetic $360^\circ$\cite{mildenhall2019local} (known as Blender dataset) includes eight complex non-Lambertian scenes finely modeled by Blender software,  and all images have a resolution of 800 $\times$ 800. 

$\textbf{Metrics}$. we use PSNR, SSIM\cite{wang2003multiscale}, and LPIPS-VGG\cite{zhang2018unreasonable} as the evaluation metrics. We also report the time needed in the training process.

Note that since the scene images are degraded, the pose estimation might be affected. We use COLMAP\cite{schoenberger2016sfm, schoenberger2016mvs} to estimate the camera poses with the degraded images.

\begin{table*}[t] 
	\centering
        \begin{tabular}{c|c|c|c|c|c}
        \toprule
           Method & PSNR$\uparrow$ & SSIM$\uparrow$ & LPIPS$\downarrow$ & RAYS$\downarrow$ & TIME$\downarrow$ \\
        \midrule
        w/o Implicit Optimization(Blender) & 23.86 & 0.848 & 0.179 & 174M & 1201s\\
        w/o Quadtree(Blender) & 24.50 & 0.859 & 0.171 & 875M & 4320s\\
        Ours(Blender) & 24.90 & 0.851 & 0.187 & 338M & 1287s\\
        \midrule
        w/o Implicit Optimization(LLFF) & 21.39 & 0.663 & 0.465 & 183M & 2704s\\
        w/o Quadtree(LLFF) & 21.61 & 0.673 & 0.462 & 729M & 6364s \\
        Ours(LLFF) & 22.01 & 0.657 & 0.472 & 386M & 3050s\\
        \bottomrule
		\end{tabular}
        \caption{Experiment results on exploring the effectiveness of Implicit Optimization and Quadtree Acceleration.}
		\label{tab:ablation}
\end{table*}

\subsection{Results}

We render at $800 \times 800$ as the target resolution for Blender dataset and $2016\times 1512$ for LLFF dataset.

Table.\ref{tab:main results} summarizes the quantitative results of our experiments. It can be seen that our proposed method outperforms the baseline method by a large margin. Due to the affection of the implicit optimization strategy, the training cost increases, and quadtree acceleration reduce such training cost greatly. 

Fig.\ref{fig:qualitative} shows the qualitative results of our experiments. It can be seen that our method can generate views of higher quality. Baseline methods suffer from severe noise and artifacts. While our method produces cleaner views of higher quality. Although our method can generate sharper details with a high-resolution training set(LLFF), the preprocessing network fails to recover the edges and texture when input images are severely degraded.

\subsection{Ablation Studies}

To verify the effectiveness of every module we propose in this paper, we conduct extensive ablation studies. Through these experiments, we can analyze how our design of restoration network, implicit optimization and quadtree acceleration work. Note that we conduct these experiments with Instant-NGP as the backend.

\paragraph{Effectiveness of Restoration Network}

\begin{figure}[h]
    \centering
    \begin{subfigure}{0.49\linewidth}
        \centering
        \includegraphics[width=\linewidth]{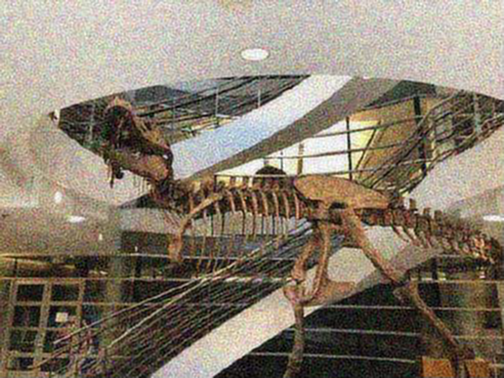}
        \label{subfig:image1}
        \subcaption{Degraded}
    \end{subfigure}
    \hfill
    \begin{subfigure}{0.49\linewidth}
        \centering
        \includegraphics[width=\linewidth]{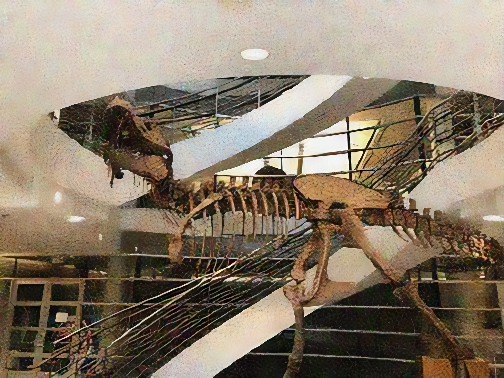}
        \label{subfig:image2}
        \subcaption{Real-ESRGAN}
    \end{subfigure}

    \vspace{0.2cm}

    \begin{subfigure}{0.49\linewidth}
        \centering
        \includegraphics[width=\linewidth]{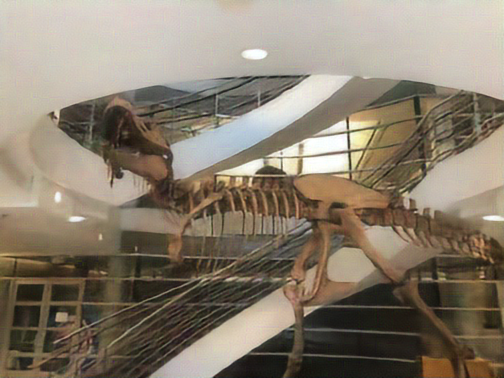}
        \label{subfig:image3}
        \subcaption{Ours}
    \end{subfigure}
    \hfill
    \begin{subfigure}{0.49\linewidth}
        \centering
        \includegraphics[width=\linewidth]{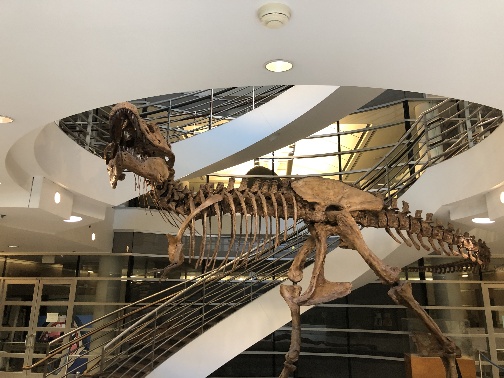}
        \label{subfig:image4}
        \subcaption{GT}
    \end{subfigure}

    \caption{Preprocessing results for trex in LLFF with different preprocessing methods.}
    \label{fig:2x2_images}
\end{figure}

As shown in Table.\ref{tab:ablation_pre_blender} and Table\ref{tab:ablation_llff}, we have tested different methods to restore the training set. See Fig.\ref{fig:2x2_images} for a detailed example. We tried not to restore the images(w/o restoring), and it leads to results of low quality. Using Real-ESRGAN, which is designed for the degradation model we utilize, yields noises on the output images. This problem is severe on LLFF dataset. This is because of the gap between the LLFF dataset and the training data of Real-ESRGAN. Then we explored only using a single image(single-view) to restore the images or removing the view selection(multi-view), which also causes a performance drop. Our method(multi-view(VS)) outperform all the other method, indicating the effectiveness of our designed preprocessing network.  

\paragraph{Effectiveness of Implicit Optimization}

As reported in Table\ref{tab:ablation}, for image quality metrics, using implicit optimization will significantly enhance the image quality. However, once adopt quadtree acceleration, although there will be an increment in PSNR, SSIM and LPIPS will drop a little. It is because applying quadtree acceleration will change the distribution of the training rays. The training will turn to focus on the edges and details.

Using implicit optimization will horribly increase the training cost. As shown in Table\ref{tab:ablation}, rays used for training increase significantly. Applying quadtree acceleration greatly reduced the horrible training cost. However, as explained before, the restoration model and NeRF model can easily deal with the low-frequency component. Even with quadtree acceleration applied, most of the rays are not sampled towards these trivial regions, the generated quality of the NeRF model is not degraded that much. On the contrast, we can train the NeRF model much faster.
\section{Limitations and Conclusions}

Our method has several limitations. First, RustNeRF does not expoit bundle adjustment to deal with the camera poses estimation offset caused by the degradations. Currently we only use COLMAP to estimate the poses again with the degraded images, which might not be accurate. Second, there might be better solutions for the degradation model when training the restoration network. Now we only combine several classical degradation models to estimate the degradation process, which could be improved. We believe that introducing better degradation model will enhance the capability of the restoration network, thus improve the generated image quality. Besides, except for implicit supervision, some explicit supervision with multi-view prior are yet to be exploited. We believe that with explicit supervision, the redundant information in multiple views can be better utilized. These will be left for future work.

In conclusion, we presented RustNeRF for robust novel view synthesis with low-quality images. We excavated the video dataset and simulated real-world degradation model and trained a general 3D-aware restoration network for these degraded training set. The restoration network can greatly restore the degraded images. We further design a method to utilize the information redundancy in different views with implicit optimization to enhance image quality. With RustNeRF, even with very low-quality input image set, the NeRF model can still generate high-quality results, getting rid of severe artifacts. RustNeRF can be applied to a various of fields, enabling future downstream applications like AR/VR. We are looking forward to step into the 3D world in the near future with so many wonderful NeRF models.

{
    \small
    \bibliographystyle{ieeenat_fullname}

    \bibliography{main}
}

\clearpage
\setcounter{page}{1}
\maketitlesupplementary

\appendix
\section{Overview}
To help readers better comprehend our method and experiments, we provide more details of RustNeRF's in the supplementary material. The supplementary content covers the following perspectives:
\begin{itemize}
    \item Degradation Parameterization
    \item Restoration Network
    \item Quadtree Acceleration
    \item Training Recipe
    \item More Qualitative Results
    \item Demo Video
\end{itemize}

\section{Degradation Parameterization}

We design our degradation space $\Omega$ with second-order degradation model described in \cite{wang2021real}. First we sharpen the image-set with USM Sharpener. In the first degradation iteration, we serially apply following operations. The kernel sizes in all the following operations are the same value in the same scene. The value is a random odd number in $[7,21]$.

\begin{enumerate}
    \item \textbf{Blur.} Kernels of \textit{iso}, \textit{aniso}, \textit{generalized iso}, \textit{generalized aniso}, \textit{plateau iso}, \textit{plateau aniso} will be randomly selected. $\sigma, \beta_g, \beta_p$ are randomly picked in range $[0.2, 3], [0.5, 4], [1, 2]$ respectively.
    \item \textbf{Resize.} We randomly use 'area', 'bilinear' or 'bicubic' mode to interpolate the image with a random scale factor in range $[0.15, 1.5]$.
    \item \textbf{Add noise.} We randomly add gray noise, gaussian noise or poisson noise.
    \item \textbf{JPEG compression.} The image quality is seleted randomly between $[30, 95]$.
\end{enumerate}

The above operations are adopted once more with newly generated random parameters.

Finally, resize, JPEG compression and sinc filter is adopted in two probable orders(both 50\%), which is resize + sinc filter + JPEG compression or JPEG compression + resize + sinc filter. In this part, the image is resized to the target resolution, with equal probabilities using 'area', 'bilinear' or 'bicubic' mode for interpolation. The image quality for JPEG compression is a random value in $[30, 95]$.

For each scene, we sample a parameter-set $\Theta$ to conduct the degradation for the whole scene.

\section{Restoration Network}

The image quality of the training set is important in training a NeRF. It is hard to reconstruct a complex scene well with degraded images of low quality. However, restoring a degraded image with an unknown and complex degradation process is a challenging task. To address this problem, we propose to train a 3D-aware restoration network to restore the degraded images before training the NeRF model.

\paragraph{Network Architecture}

Following NeRFLiX, first we use a simple encoder to extract the features of different views. The encoder network first output features with two convolution layers that downscale the features by 1/2 and two convolution layers followed by two residual blocks. Then three different scales of features are calcuated by a LeakyReLU, one convolution layer with stride 2 followed by 2 residual blocks and two convolution layers with stride 2 followed by 7 residual blocks respectively. Then features of all scales are concatenated into the final feature.

After extracting the features of all the views, we can conduct the $f_{res}$ process described in our paper. We can aggregate all the feature before restoring the target view. We use the Hybrid Recurrent Aggregation described in NeRFLiX to recurrently aggregate the information from different views to the target view, then use a simple CNN to reconstruct the target view.

\paragraph{Network Training}

We leverage the characteristics of the video dataset, as consecutive frames in the video dataset are naturally multiple views of a scene, which is similar to the dataset used in NeRF.

Given $n$ frames of a scene, i.e. a clip in the video dataset, denoted as $\mathit{I}=\{I_1,I_2,...,I_n\}$, we first adopt scene degradation through degradation parameterization and get the degraded frames $\mathit{I'}=\{I'_1,I'_2,...,I'_n\}$. At the training stage, we randomly select three different degraded frames $I'_i, I'_j, I'_k$ and the corresponding original frame $I_i$ to construct paired data. We merge the training part of LLFF and Vimeo90K \cite{xue2019video} as the training dataset.

\section{Quadtree Acceleration}

\begin{figure}[h]
    \begin{subfigure}{0.49\columnwidth}
        \centering
        \includegraphics[width=\linewidth]{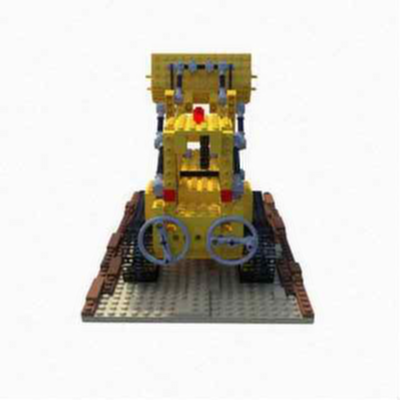}
        \caption{}
        \label{fig:subfig1}
    \end{subfigure}
    \begin{subfigure}{0.49\columnwidth}
        \centering
        \includegraphics[width=\linewidth]{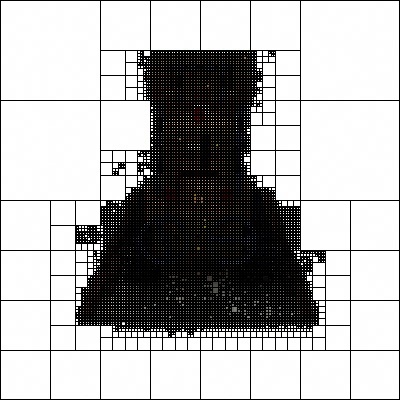}
        \caption{}
        \label{fig:subfig2}
    \end{subfigure}
    \caption{Example of subdividing a quadtree. (a) is the original training image. (b) is the visualization of the quadtree after training. The leaf nodes on the trivial part of the image stops splitting early. Less rays are sampled in these parts, which helps to reduce the amounts of rays used in training.}
    \label{fig:bothfigs}
\end{figure}

Implicit multi-view optimization strategy significantly increases the number of required rays during training by a factor of $s^2$, which is not feasible. 

The previous work\cite{zhang2023fast} explored how to use quadtree to accelerate the training of NeRF. Inspired by this, we design a tactful strategy to determine supersampling pixels based on quadtree to reduce the training cost.

Intuitively, NeRF model can easily learn low-frequency information, which refers to areas with similar colors in larger areas. In addtion,  while it is more difficult for the model to learn high-frequency information, which includes edges and details. 

Traditional uniform sampling method samples each pixel with the same probability. Our method assigns different probabilities to each super-pixel, based on how much it might contribute to the image quality. We aim to sample more frequently the super-pixels covering the blurred edges or other details, while reducing the sampling of regions that are relatively simple and trivial.

Based on this prior knowledge, we design a super-pixel sampler based on quadtree. Through assigning higher probabilities to the super-pixels that are more important, we accelerate the convergence of training, and reduce the number of rays required for training. 

In the traditional method, the probability of each super-pixel to be sampled each epoch is equal. After introducing the sampler, we reassign the sampling probability to each super-pixel in the training set $\mathcal{R}$ according to the information of the quadtree and the ground truth images before each epoch of training. Let $\mathcal{D}$ be the set consisting of all leaf nodes of the quadtree. For each leaf node $D\in\mathcal{D}$, the sampling probability of a super-pixel $p\in D$ is calculated based on the variance of its RGB values with its eight neighboring pixels, and then normalized within that region.

\begin{equation}
g(p(i, j))=\sqrt{\frac{1}{9} \sum_{(x, y)\in (i,j)\cup N_8(i, j)}[C(x, y)-\overline{C}]^2}
\end{equation}

\begin{equation}
    \mathcal{P}(p(i, j))=\frac{g(p(i, j))}{\max_{p_n\in D} (g(p_n(i, j)))}
\end{equation}
where $\mathcal{P}(p(i,j))$ represents the probability of sampling the super-pixel, and $N_8(i, j)$ represents the eight neighborhoods of super-pixel $(i, j)$.

After redistributing the sampling probabilities, we construct the training set $\mathcal{R}_e$ for the training in a new epoch of $e$ according to the following strategy. The hyper-parameter $\mu$ represents sampling density. For each leaf node $D$ of all current quadtrees, the number of rays sampled within it is given by

\begin{equation}
    n_D= 
    \begin{cases}
    \mu \times \operatorname{area}(D), & \text { if } \mathcal{L}^{e-1}_{MSE}(D) < s_{sample} \\ 
    \alpha \times \mu \times \operatorname{area}(D), & \text { otherwise }
    \end{cases}
\end{equation}
Where $s_{sample}$ represents the threshold used in sampling and $\alpha$ is a hyper-parameter controlling the sampled density in these well-learned areas. Intuitively, we set $\alpha=0.1$.

For each leaf node $D$, we can calculate the number of super-pixels to be sampled. By sampling according to the calculated probability distribution, we can obtain a new training set $\mathcal{R}_e$ for the next epoch. Finally, during training, we adopt the uniform sampling on training set $\mathcal{R}_e$ to achieve our goal, i.e. sample more frequently those important super-pixels.

The subdivision operation may only be performed to the deepest level leaf nodes. During the subdivision process, for each deepest leaf node $D'$, we calculate the average loss values of all super-pixels sampled in $D'$ in the previous epoch. If this average value exceeds the predefined threshold $s_{divide}$, the node is subdivided, otherwise, the node is skipped. Specifically, the subdivision operation refers to dividing the area covered by $D'$ into four sub-nodes:  top-left, bottom-left, top-right, and bottom-right, denoted as $D'_1, D'_2, D'_3, D'_4$. These four nodes are added to the leaf node set $\mathcal{D}$, while $D'$ is removed from the set.

We subdivide the quadtrees after the training of each epoch as we find it helps the quadtrees converge faster. We set the threshold for subdivision $s_{divide}=0.02$. We randomly sample 20\% of the rays with normal strategies to enhance the generalizability. Before the training starts, we fisrt subdivide the quadtrees twice as we found that quadtrees with too few nodes do not contribute to the training process. To prevent the quadtrees from being subdivided into too many nodes, which will greatly increase the overhead of relevant operations of quadtrees, we do not subdivide the nodes whose area is less or equal to $625$.

\begin{figure*}
    \includegraphics[width=\linewidth]{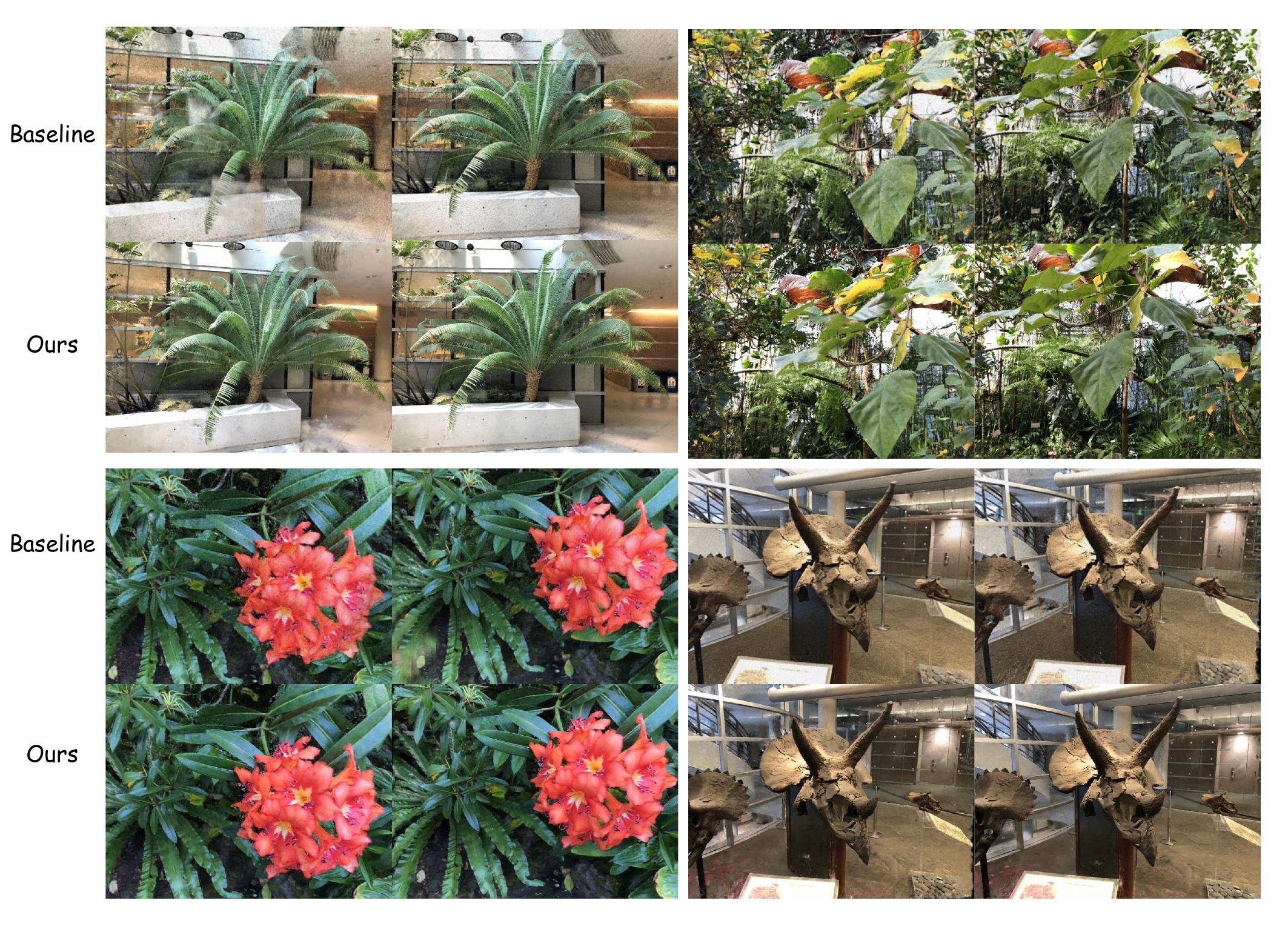}
    \caption{We display more experimental results to showcase our qualitative results on LLFF dataset.}
    \label{fig:more_results_llff}
\end{figure*}

\section{Training Recipe}
\subsection{Ours(DVGO)}

The initial learning rate is 1e-1 for density voxel grid, 1e-1 for color voxel grid and 1e-3 for the MLP that predicts the color. Adam is used as the optimizer and $\beta_1, \beta_2$ are set to 0.9 and 0.999 respectively.

For Blender dataset, we use the batch size of 8192. We train for 5k iterations for the coarse geometry search. Then in the fine grain optimization process, we train for 5k iterations with normal sampling strategy and 15k iterations with supersampling strategy.

For LLFF dataset, we set the batch size to 4096. We skip the coarse geometry searching and adopt fine grain optimization directly. We train for 5k iterations with normal sampling and 25k iterations with supersampling.

\subsection{Ours(Instant-NGP)}

For Blender dataset, we use 2e-2 as the initial learning rate. We set the batch size to 16384. We train with normal sampling strategy for 4 epochs and then train with supersampling strategy for 12 epochs.

For LLFF dataset, we use 1e-2 as the initial learning rate. Batch size is set to 8192. We train for 3 epochs with normal strategy and 6 epochs with supersampling.

We use Adam as the optimizer, whose $\beta_1, \beta_2$ are set to 0.9 and 0.999. We adopt cosine annealing strategy for the learning rate. The learning rate is decayed to 1/10 at the end of training. 

\subsection{More Qualitative Results}
\begin{figure*}[ht]
    \includegraphics[width=\linewidth]{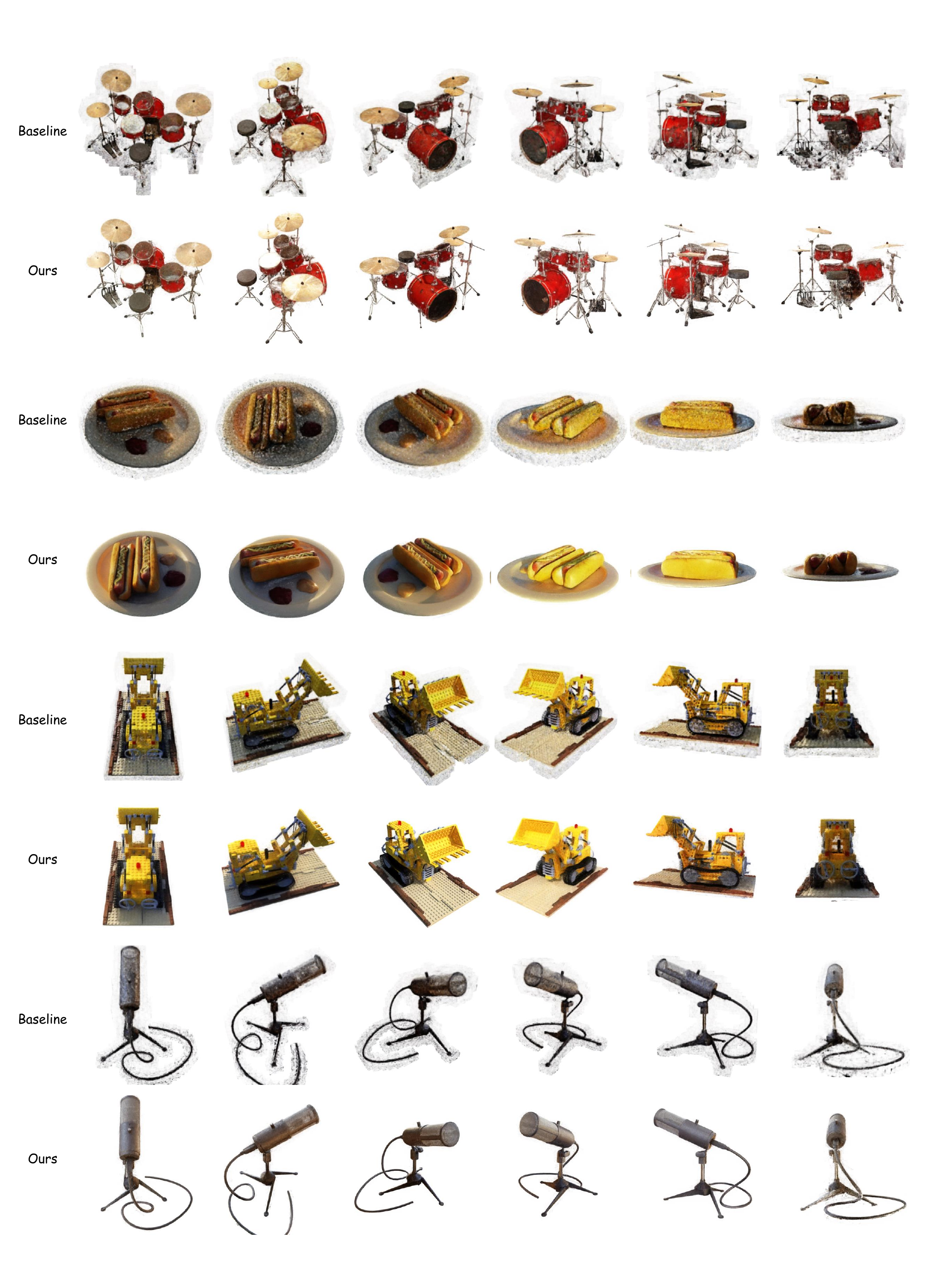}
    \caption{We display more experimental results to showcase our qualitative results on Blender dataset.}
    \label{fig:more_results_blender}
\end{figure*}

We offer additional visual examples to effectively demonstrate the success of our method. As seen in Fig.7 and Fig.8, RustNeRF consistently improves the rendered images by providing clearer details and reducing artifacts. For instance, RustNeRF effectively recovers recognizable characters, object textures, and realistic reflectance effects while eliminating rendering artifacts. We also supply a video demo for easy visual comparison. We display some scenes rendered with baseline and their RustNeRF enhanced counterparts.

\end{document}